\documentclass[letterpaper, 10 pt, conference]{ieeeconf}  % Comment this line out if you need a4paper

\IEEEoverridecommandlockouts                              % This command is only needed if 
\usepackage{graphics} % for pdf, bitmapped graphics files
\usepackage{graphicx}
\usepackage{epsfig} % for postscript graphics files
\usepackage{times} % assumes new font selection scheme installed
\usepackage{amsmath} % assumes amsmath package installed
\usepackage{amssymb}  % assumes amsmath package installed
\usepackage{bbm}
\usepackage{hyperref}
\usepackage{csquotes}
\usepackage{cite}
\usepackage{booktabs}
\usepackage{siunitx}

\usepackage[utf8]{inputenc}
\usepackage[english]{babel}
\newcommand{\norm}[1]{\left\lVert#1\right\rVert}

\title{\LARGE \bf
Neural Implicit Vision-Language Feature Fields
}

\author{Kenneth Blomqvist$^{1}$, Francesco Milano$^{1}$, Jen Jen Chung$^{2}$, Lionel Ott$^{1}$ and Roland Siegwart$^{1}$% <-this % stops a space
\thanks{$^{1}$Autonomous Systems Lab, Swiss Federal Institute of Technology in Z\"urich, Switzerland. 
    {\tt\small kblomqvist@mavt.ethz.ch}}%
\thanks{$^{2}$School of ITEE, The University of Queensland, Australia.}%
\thanks{This project has received funding from EU Horizon 2020 program, project PILOTING H2020-ICT-2019-2 871542.}
}

\begin{document}

\maketitle
\thispagestyle{empty}
\pagestyle{empty}

%%%%%%%%%%%%%%%%%%%%%%%%%%%%%%%%%%%%%%%%%%%%%%%%%%%%%%%%%%%%%%%%%%%%%%%%%%%%%%%%
\begin{abstract}

Recently, groundbreaking results have been presented on open-vocabulary semantic image segmentation. Such methods segment each pixel in an image into arbitrary categories provided at run-time in the form of text prompts, as opposed to a fixed set of classes defined at training time. In this work, we present a zero-shot \emph{volumetric} open-vocabulary semantic scene segmentation method. Our method builds on the insight that we can fuse image features from a vision-language model into a neural implicit representation. We show that the resulting feature field can be segmented into different classes by assigning points to natural language text prompts. The implicit volumetric representation enables us to segment the scene both in 3D and 2D by rendering feature maps from any given viewpoint of the scene. We show that our method works on noisy real-world data and can run in real-time on live sensor data dynamically adjusting to text prompts. We also present quantitative comparisons on the ScanNet dataset.
\end{abstract}

%%%%%%%%%%%%%%%%%%%%%%%%%%%%%%%%%%%%%%%%%%%%%%%%%%%%%%%%%%%%%%%%%%%%%%%%%%%%%%%%
\section{INTRODUCTION}

A key component of building intelligent robots capable of operating in unstructured and cluttered human environments is the representation used to model the robot's surroundings.
Often times representations have to trade-off properties which depend on the usage scenario. These properties include the quality of the reconstruction, the ability to integrate sensor data continuously, and the computational complexity to query the representation. The importance of these aspects differs based on what components of a robotic system needs to use the representation, dictating the requirements for available capabilities, sensor data throughput, or query latency. For instance, an obstacle avoidance system needs to query for occupancy at high frequency, while a high-level planning system needs access to semantic knowledge, and finally a grasp planning system requires fine-grained segmentation information.

While in the past occupancy was the main information of interest, robotics has moved towards richer representations using semantics in recent years. A challenge is that most semantic approaches use a fixed, closed set, of pre-determined semantic labels. However, real environments contain more than a few dozen classes, and thus methods capable of handling arbitrary semantic classes, i.e. open set, are desirable. Additionally, objects in an environment do not necessarily belong to distinct, mutually exclusive classes. Certain objects might belong to several classes. A bookshelf is also a piece of furniture, for example. For high-level planning purposes, being able to reason about relations between their semantics might also be useful.

An environment representation that has wide applicability has several desirable properties, including: (1) can be built incrementally as the robot explores the environment, (2) enables real-time integration of new measurements, (3) has a compact memory footprint, (4) represents geometry at a high-level of detail, (5) is differentiable, (6) supports open set semantic queries, and (7) allows fast querying by downstream modules. Previously introduced 3D semantic scene representations are either built from global scene information \cite{peng2022openscene}, use closed set semantics \cite{grinvald2019volumetric, rosinol2020kimera, zhi2021ilabel, mazur2022feature}, operate on a fixed level of detail \cite{grinvald2019volumetric, rosinol2020kimera}, or are not differentiable \cite{grinvald2019volumetric, rosinol2020kimera}. In this paper, we take a step towards a representation which has the above-mentioned properties.

\begin{figure}[t]
    \centering
    \includegraphics[width=0.8\linewidth]{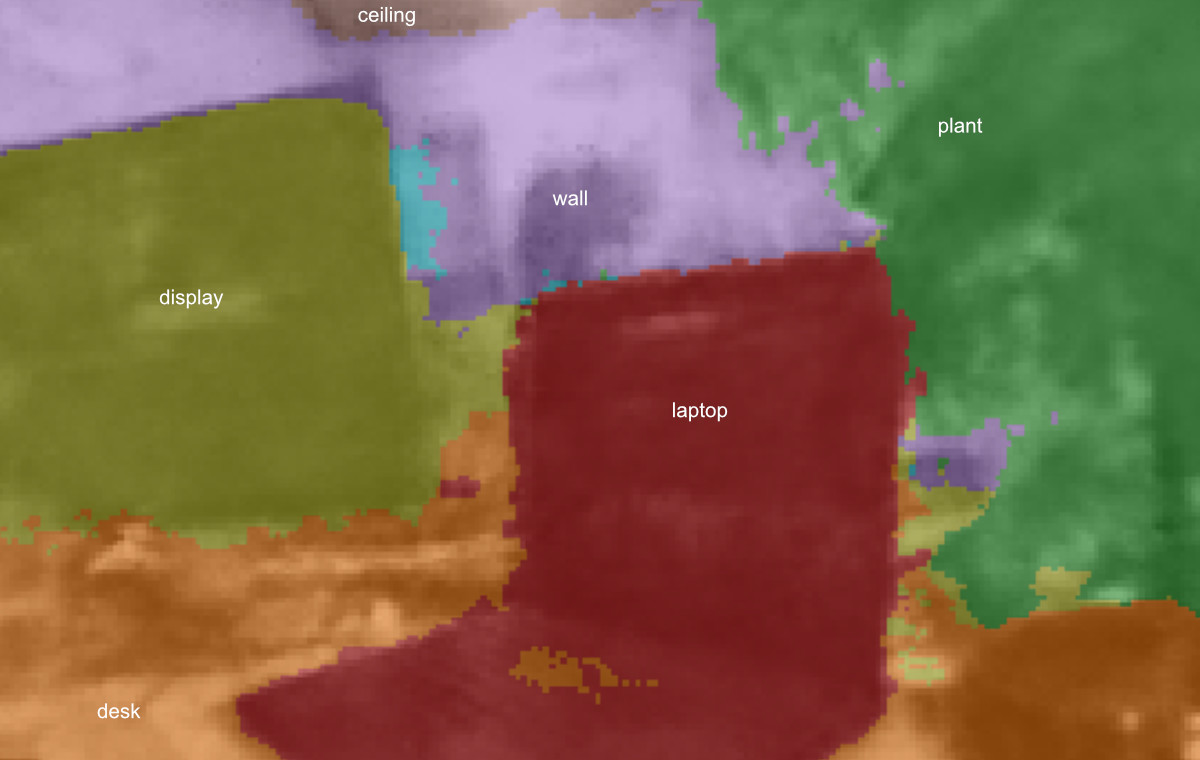}
    \caption{Our method enables real-time segmentation of scenes into arbitrary text classes provided at run-time.}
    \label{fig:main}
    \vspace{-0.2cm}
\end{figure}

Vision-language models (VLM) have shown remarkable performance on open vocabulary object detection \cite{zareian2021open, gu2021open}. Recently, these results have been extended to dense semantic segmentation \cite{ghiasi2022scaling, li2022language, zhang2022glipv2, zou2022generalized}. Some of these methods \cite{ghiasi2022scaling, li2022language} associate each pixel with a semantically meaningful vector, which is embedded in the same high-dimensional vector space as natural language prompts through a text encoder. This allows direct computation of the similarity between text prompts and image features at run-time.

As vision-language models can be trained on massive web-scale datasets that can be collected automatically without human supervision, they often show better generalization capabilities than models trained on smaller closed-set manually curated datasets. Additionally, VLMs can capture the long tail of scenarios and classes that are so rare that they are unlikely to be included in curated datasets. These properties offer great promise for applications in robotics, where we might want our robots to be able to perform new tasks in never-before-seen environments. 

In this paper, we present a method for grounding dense vision-language features into a 3D implicit neural representation that can be built up incrementally, in real-time, as new observations come in. We jointly model radiance, vision-language model features, and density in the scene using an implicit neural representation. Our representation can be incrementally built up given posed images of the scene and a pre-trained language model. We can directly compute the similarity between natural language text prompts and either 3D points or 2D image coordinates for any given viewpoint of the scene through volumetric rendering. This enables semantically segmenting a scene zero-shot into text categories provided at run-time, without having to fine-tune the system on any domain specific semantics.

In experiments, we showcase results in real-world experiments where we build up our scene representation in real-time on a real system, and demonstrate the ability to segment the scene into different classes provided as natural language prompts at run-time. We additionally present quantitative segmentation results on the large and diverse ScanNet dataset. To the best of our knowledge, our method is the first real-time capable 3D vision-language neural implicit representation. Our implementation will be made available through the Autolabel project \footnote{https://github.com/ethz-asl/autolabel}.

% Need to check if the fused features are actually better than the raw 2D features. If that is the case, this would be a nice result to showcase.

\section{RELATED WORK}

\subsection*{Open Vocabulary Semantic Segmentation and Vision-Language Models}

CLIP \cite{radford2021learning} introduced a visual-language model capable of mapping images into the same vector space as natural language queries by correlating images to their text descriptions mined from the open web.
Open vocabulary segmentation methods typically learn dense features which are compared to text queries given at run-time \cite{li2022language, ghiasi2022scaling}. Others take a multi-task learning approach, fusing a task prompt with the architecture \cite{zhang2022glipv2, zou2022generalized}.
Other methods such as Clippy \cite{ranasinghe2022perceptual} explored learning pixel-aligned visual-language models from large scale web datasets without requiring segmentation labels, potentially enabling large-scale open set training, if the results can be extended to full semantic segmentation. 

\subsection*{Language Models in Robotics}

Large language models have been explored as an approach to high-level planning \cite{ahn2022can, huang2022visual, song2022llm, chen2022open, raman2022planning} and scene understanding \cite{chen2022leveraging, ha2022semantic}. Vision-language models embedding image features into the same space as text have been applied to open vocabulary object detection \cite{song2022llm, chen2022open}, natural language maps \cite{blukis2021few, chen2022open, shafiullah2022clip, huang2022visual, tan2022self}, and for language-informed navigation \cite{shah2022lm, wang2022find, majumdar2022zson}. 

Recent methods have explored fusing global CLIP features \cite{shafiullah2022clip}, image caption embeddings \cite{ding2022language}, or dense pixel-aligned \cite{peng2022openscene} visual-language model features into a point cloud representation for scene understanding. Concurrent work ConceptFusion \cite{jatavallabhula2023conceptfusion} explores building multi-modal semantic maps by fusing features from vision-language models as well as audio into a reconstructed 3D point cloud. Similar to these, we also fuse VLM features into a 3D representation. Unlike \cite{shafiullah2022clip, peng2022openscene, jatavallabhula2023conceptfusion}, we use a continuous neural representation of geometry and semantics which we learn jointly through volumetric rendering. \cite{peng2022openscene, ding2022language} fuse image features from a pre-built point cloud using a multi-view fusion method and learn a 3D convolutional network to map scene points to dense features. Our representation can be built incrementally as measurements are collected and does not require global scene geometry upfront.

\subsection*{Semantic Scene Representations}

Voxel-based map representations have been proposed to store semantic information about a scene \cite{strecke2019fusion, grinvald2019volumetric, narita2019panopticfusion, rosinol2020kimera, schmid2022panoptic}. These methods assign a semantic class to each individual voxel in the scene. Voxel-based dense semantic representations typically operate on static scenes, but some have explored modeling dynamic objects \cite{xu2019mid, grinvald2021tsdf++}. 

Scene graphs \cite{armeni20193d, hughes2022hydra, wu2021scenegraphfusion} have also been proposed as a candidate for a semantic scene representation that can be built-up online. Such methods decompose the scene into a graph where edges model relations between parts of the scene. The geometry of the parts are typically represented as a signed distance functions stored in a voxel grid \cite{huang2022visual}.

Neural implicit representations infer scene semantics \cite{zhi2021place, zhi2021ilabel, blomqvist2022baking, mazur2022feature, fu2022panoptic, siddiqui2022panoptic, liu2022unsupervised, kundu2022panoptic} jointly with geometry using a multi-layer perceptron or similar parametric model. These have been extended to dynamic scenes \cite{kong2023vmap}. Neural feature fields \cite{kobayashi2022d3f, tschernezkineural, blomqvist2022baking, mazur2022feature} are neural implicit representations which map continuous 3D coordinates to vector-valued features. Such representations have shown remarkable ability at scene segmentation and editing. \cite{kobayashi2022d3f} also presented some initial results on combining feature fields with vision-language features, motivating their use for language driven semantic segmentation and scene composition.

\section{METHOD}

Our method consists of two components: i) a NeRF-like feature field mapping points in a volume to color, density, and feature vector and ii) a vision-language model which both extracts features from image frames and can embed text prompts into the same vector space. 

\subsection{Volumetric Scene Representation}

\begin{figure}[bt]
    \centering
    \includegraphics[width=0.7\linewidth]{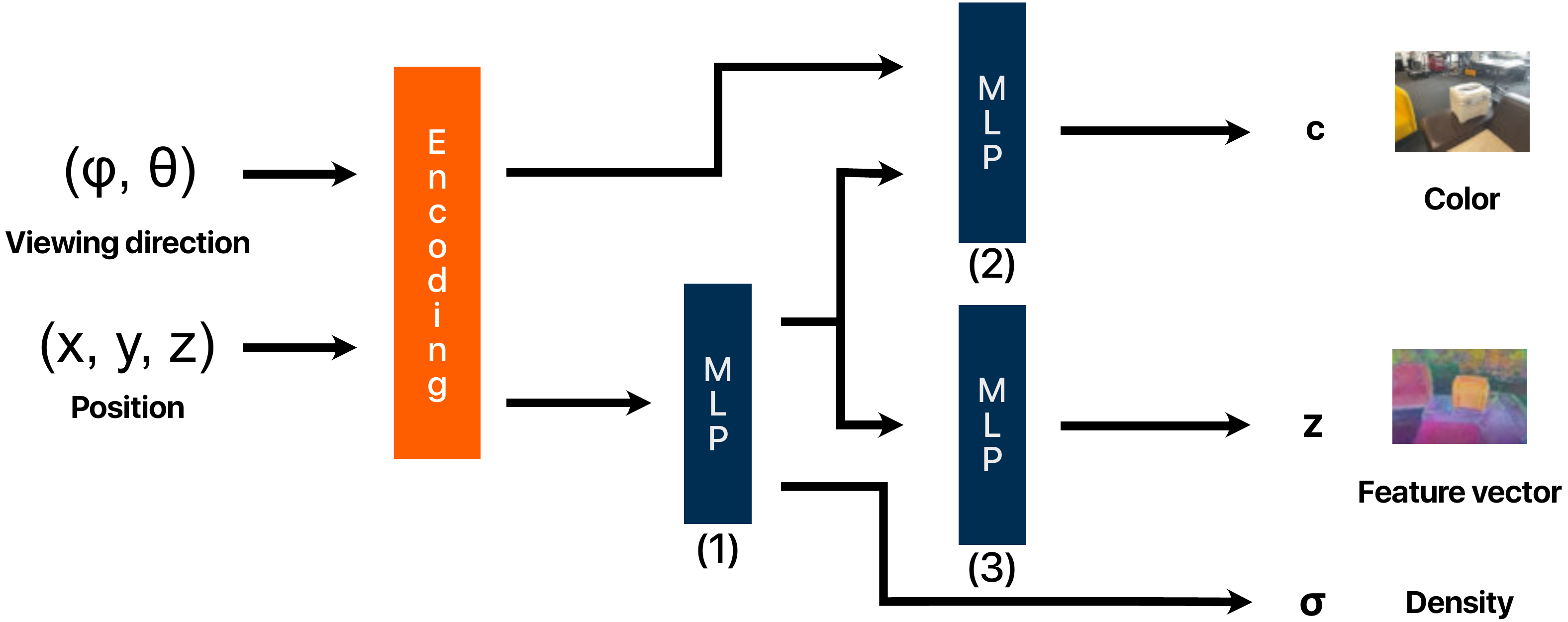}
    \caption{A diagram of the model used for our feature field.}
    \label{fig:diagram}
    \vspace{-1.5em}
\end{figure}

We want to associate 3D points in the volume of our scene to density, color, and a feature vector. From this, we can render corresponding maps of color, depth, and feature vectors through a NeRF-like \cite{mildenhall2020nerf} volumetric rendering function, visualized in Figure \ref{fig:diagram}. We model these maps using a positional encoding function and three multilayer perceptrons (MLP). The first MLP, indicated as (1), outputs density and a geometric code. The second MLP, labeled (2), outputs color from the geometric code and an encoded viewing direction. The third MLP, denoted by (3), takes the geometric code and outputs the feature vector.

To encode the $x, y$, and $z$ position in the volume, we use the hybrid positional encoding introduced by \cite{blomqvist2022baking}. We concatenate the vector valued hashgrid encoding introduced in \cite{muller2022instant} with the low-frequency values of traditional NeRF \cite{mildenhall2020nerf} frequency encoding with $L=2$. The low-frequency components allows us to model the coarse spatial location in the scene, whereas the parameters in the hashgrid grid allow us to quickly learn high-frequency details.

The resulting encoding is fed into an MLP, (1) in Figure \ref{fig:diagram}, which outputs a $15$-dimensional geometric code vector and scalar density $\sigma$. The geometric code is fed into two different MLPs. The first one outputs a feature vector $\boldsymbol{f}$. The other one takes as additional input the encoded viewing direction and outputs a color vector $\boldsymbol{c}$. To encode the viewing direction, we use the same spherical harmonic encoding as \cite{mildenhall2020nerf}.

We use these outputs to volumetrically render color images and feature outputs using the rendering function:
\begin{align}
\label{eq:render}
R(\mathbf{r}, h) &= \sum_{i=1}^N T_i (1 - \exp(-\sigma_i \delta_i)) h(\mathbf{x}_i),\\
T_i &= \exp\left(-\sum_{j=1}^{i-1} \sigma_j \delta_j\right),
\end{align}
where $h$ is a function outputting a vector or scalar quantity for points $\mathbf{x}_i$ within the volume, $T_i$ is the transmittance function, $\delta_j$ is the distance between samples and $\sigma_i$ is the predicted density for encoded point samples $\mathbf{x}_i$ along a ray $\mathbf{r}$. We use $R$ to produce rendered quantities:
\begin{align}
\begin{split}
\label{eq:maps}
\hat{\mathbf{c}}(\mathbf{r}) &= R(\mathbf{r}, \mathbf{c}), \\
\hat{d}(\mathbf{r}) &= R(\mathbf{r}, z), \\
\hat{\mathbf{f}}(\mathbf{r}) &= R(\mathbf{r}, \mathbf{f}),
\end{split}
\end{align}
using $z$ for the depth component of samples, $\mathbf{c}$ for the color MLP output and $\mathbf{f}$ for the feature vector output of our MLP.

These quantities are learned by optimizing photometric, depth, and feature rendering error terms:
\begin{align}
\mathcal{L}_{rgb}(\mathbf{r}) &= \norm{\hat{\mathbf{c}}(\mathbf{r}) - \bar{\mathbf{c}}(\mathbf{r})}^2_2, \\
\mathcal{L}_{d}(\mathbf{r}) &= \begin{cases}
\norm{{\hat{d}(\mathbf{r})} - \bar{d}(\mathbf{r})}_1, & \text{if $\bar{d}$ is defined for $\mathbf{r}$} \\
0, & \text{otherwise}
\end{cases} \\
\mathcal{L}_{f}(\mathbf{r}) &= \norm{\hat{\mathbf{f}}(\mathbf{r}) - \bar{\mathbf{f}}(\mathbf{r})}^2_2 / D
\end{align}
where $\bar{\mathbf{c}}(\mathbf{r})$ is the ground truth and $\hat{\mathbf{c}}(\mathbf{r})$ the predicted color for ray $\textbf{r}$, $\bar{d}(\mathbf{r})$ is the ground truth depth (if available), $\hat{d}(\mathbf{r})$ the predicted depth predictions along ray $\mathbf{r}$, $\hat{\mathbf{f}}$ rendered feature outputs, $\bar{\mathbf{f}}$ extracted image features for ray $\mathbf{r}$, and $D$ the dimensionality of the image features.

The parameters in the hashgrid encoding volume and in the MLPs are jointly learned by optimizing the objective:
\begin{equation}
\mathcal{L}(\mathbf{r}) = \mathcal{L}_{rgb}(\mathbf{r}) + \lambda_d \mathcal{L}_{d}(\mathbf{r}) + \lambda_f \mathcal{L}_{f}(\mathbf{r})
\end{equation}
using stochastic gradient descent on a set of rays sampled uniformly from input images $\mathbf{I}$ along with corresponding feature vectors $\bar{\mathbf{f}}$. The parameters $\lambda_d$ and $\lambda_f$ are weighting parameters to weight the different components of the loss function. To learn the representation online, while our robot is exploring the environment, keyframes with image features can be added to the image set as they are captured.

\subsection{Vision-language Features and Zero-shot Segmentation}

Our framework presented above is capable of making use of arbitrary feature maps. Thus, we can use features from any feature extractor that produces dense pixel-aligned feature maps from images. To enable open set semantic queries in both 2D and 3D at run-time, we choose to use learned features for which the similarity with text prompts can be computed through a simple dot product. LSeg \cite{li2022language} and OpenSeg \cite{ghiasi2022scaling} are both suitable candidates for this purpose. 
% XDecoder \cite{zou2022generalized} and similar open vocabulary segmentation methods might be an option, but they would require learning a text and image feature fusion module.
In our experiments, we use LSeg features, as pretrained models are readily available. The model comes both with an image feature extractor $\bar{\mathbf{F}}$ and text encoder $\mathbf{E}$.

Given a pose in the world frame of the volume, we can render color, depth, and feature maps using volumetric rendering, using equations \ref{eq:render} and \ref{eq:maps}. We compute the semantic class by assigning the feature $\hat{\mathbf{f}}$ to the most similar class given a set of user defined natural language class descriptions $t_i \in \mathcal{T}$ into which we want to segment our scene:
\begin{equation}
    \hat{s}(\mathbf{r}) = \text{argmax}_i \mathbf{E}(t_i) \cdot \hat{\mathbf{f}}(\mathbf{r}).
\end{equation}

For 3D queries at point $\mathbf{x}$, we can simply evaluate the feature MLP at $\mathbf{x}$, i.e.:
\begin{equation}
    s(\mathbf{x}) = \text{argmax}_i \mathbf{E}(t_i) \cdot \mathbf{f}(\mathbf{x}).
\end{equation}

\section{EXPERIMENTAL RESULTS}
\begin{figure*}[t]
\centering
\includegraphics[width=0.9\linewidth]{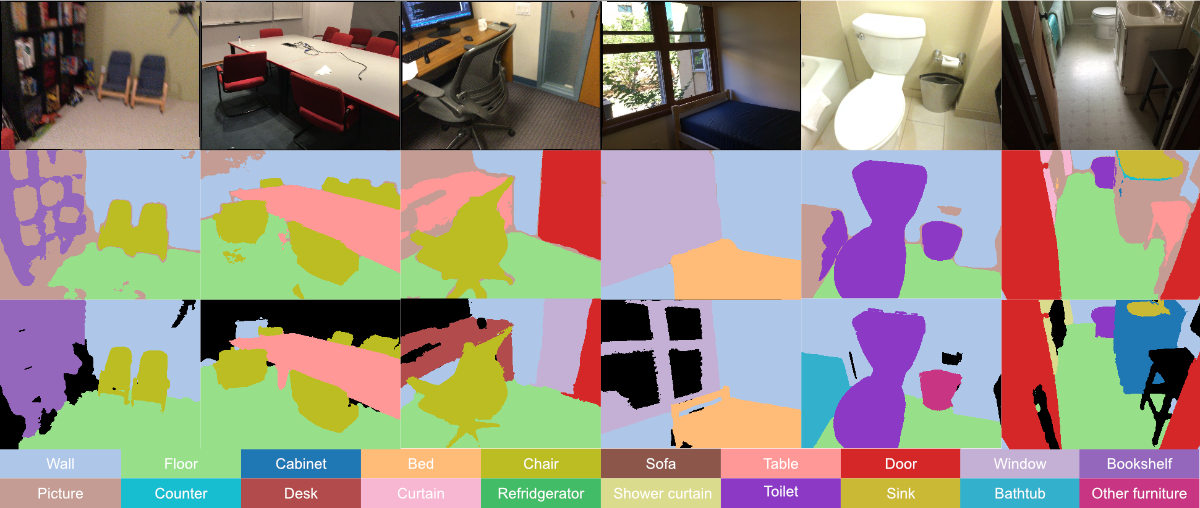}
\caption{Randomly sampled 2D segmentation examples from the ScanNet validation set. Top row shows the original RGB images, second row shows our segmentation and the bottom row shows the ground truth segmentation from the ScanNet dataset. Black pixels in the ground truth segmentation correspond to classes not included in the 20 ScanNet evaluation classes.}
\label{fig:scannet}
\end{figure*}

In the following experiments we provide quantitative results on the ScanNet dataset. We compare our method to the OpenScene \cite{peng2022openscene} work in terms of mean intersection over union ($\mathrm{mIoU}$) and mean accuracy ($\mathrm{mAcc}$). Then, to highlight the utility of our approach in robotics application we integrate our approach with a SLAM framework. Finally, we report run-time information to demonstrate the feasibility of running our algorithm on a real robotic system.

In all our experiments, we use LSeg features \cite{li2022language} trained on the ADE20k dataset \cite{zhou2017scene}. For the loss function, we use $\lambda_d = 0.1$ and $\lambda_f = 0.5$ throughout all experiments. Having tried a range of different values, we found that they perform similarly and settled on these values in the middle of the range. In case less noisy and more accurate depth measurements are available, a higher $\lambda_d$ value might yield better results.

\subsection{ScanNet}

On the ScanNet dataset we perform evaluation both in 3D, by segmenting the provided ground truth point cloud, as well as in 2D by comparing our rendered segmentation maps to the ones provided in the dataset. We use the $20$ classes from the ScanNet benchmark. Points or pixels that do not belong to these classes are ignored.

We first fit our representation using the given RGB, depth frames and camera poses using $\num{20 000}$ optimization iterations. For 3D point cloud segmentation, we look up the feature vector for each point in the point cloud and assign it to the nearest text class using the ScanNet class label names as the text prompts. For 2D segmentation, we segment feature maps from each viewpoint in each scan and compare against the reference segmentation map.

\begin{table}[h]
\centering
\begin{tabular}{l c c}
\toprule
                         & ScanNet $\mathrm{mIoU}$ & ScanNet $\mathrm{mAcc}$\\ \midrule
OpenScene - LSeg (3D)    & 54.2         & 66.6 \\
OpenScene - OpenSeg (3D) & 47.5         & 70.7 \\
Ours - LSeg (3D)         & 47.4         & 55.8 \\ 
Ours - LSeg (2D)         & 62.5         & 80.2 \\ \bottomrule
\end{tabular}
\caption{Mean intersection-over-union agreement with the ScanNet validation set.}
\label{table:scannet-overall}
\end{table}

Table \ref{table:scannet-overall} shows mean intersection-over-union ($\mathrm{mIoU}$) results on the ScanNet validation set, averaging over scenes and classes. LSeg\cite{li2022language}/OpenSeg \cite{ghiasi2022scaling} denotes the 2D image features used. 3D denotes segmentation agreement on the given ground truth point cloud whereas 2D shows agreement against the semantic segmentation maps.

OpenScene \cite{peng2022openscene} performs better overall, but it should be noted that it makes use of the ground truth scene point cloud, whereas we only use the color and depth frames and implicitly reconstruct the geometry. We only use the scene point cloud for evaluation. We additionally show 2D segmentation results compared with the ground truth segmentation frames in the dataset. As OpenScene only segments the point cloud, only 3D segmentation accuracy is shown.

Figure \ref{fig:scannet} shows qualitative 2D segmentation masks. Our method mostly performs well, but often struggles to distinguish between semantically similar classes such as ``desk" and ``table" or ``curtain" and ``shower curtain" in the ScanNet evaluation, as we do not make use of any tuning to align the semantics of the dataset with the semantics of the vision-language vector space. The ScanNet label quality is also not perfect and our method often gets details correct which are missed by the ScanNet ground-truth labels, such as legs of tables and chairs or other thin structures.

\subsection{Real-time SLAM Experiment}

\begin{figure*}[t]
    \centering
    \includegraphics[width=0.8\linewidth]{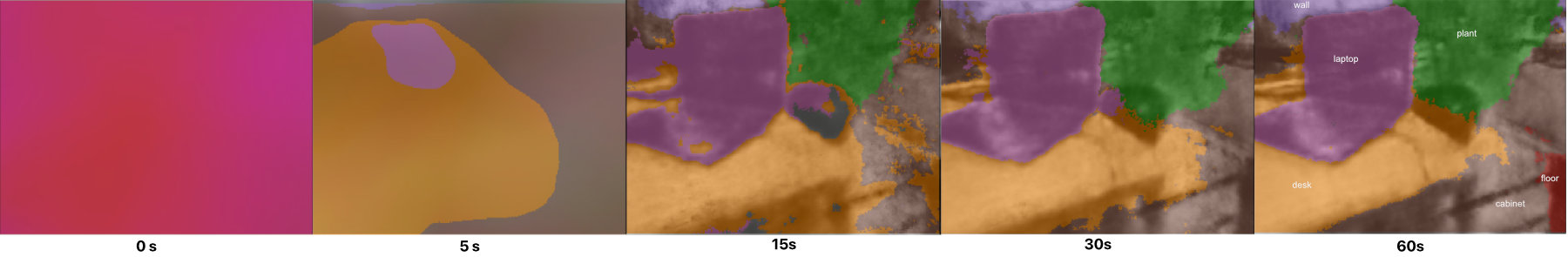}
    \vspace{-0.2cm}
    \caption{Snapshots from real-time zero shot volumetric segmentations from a fixed viewpoint at given intervals. Our representation is able to learn in real-time and is already useful after a dozen seconds. Each image shows RGB rendering output for the viewpoint, overlayed with the semantic segmentation given the $6$ class prompts shown.}
    \label{fig:timeline}
\end{figure*}

To test our scene representation in a real-world robotics scenario, we integrate our system with a SLAM pipeline \footnote{Specifically the SpectacularAI SDK available here: https://github.com/SpectacularAI/sdk} using a Luxonis OAK-D Pro stereo camera. While the system is running, we integrate color, depth, and features extracted using LSeg from keyframes at $\SI{5}{Hz}$ with poses obtained from the SLAM system. In experiments, we use either the left (grayscale) camera image or RGB camera. Depth is computed using stereo matching and aligned to the keyframe camera's frame. 

To test our system we give it classes in the form of text prompts while it is running and inspect the quality of the segmentation. Using the odometry poses provided by the SLAM system, we render color, depth maps and segmentation maps from the current camera viewpoint in real-time, segmenting the camera image into the given classes. 

\begin{figure}
    \centering
    \includegraphics[width=0.8\linewidth]{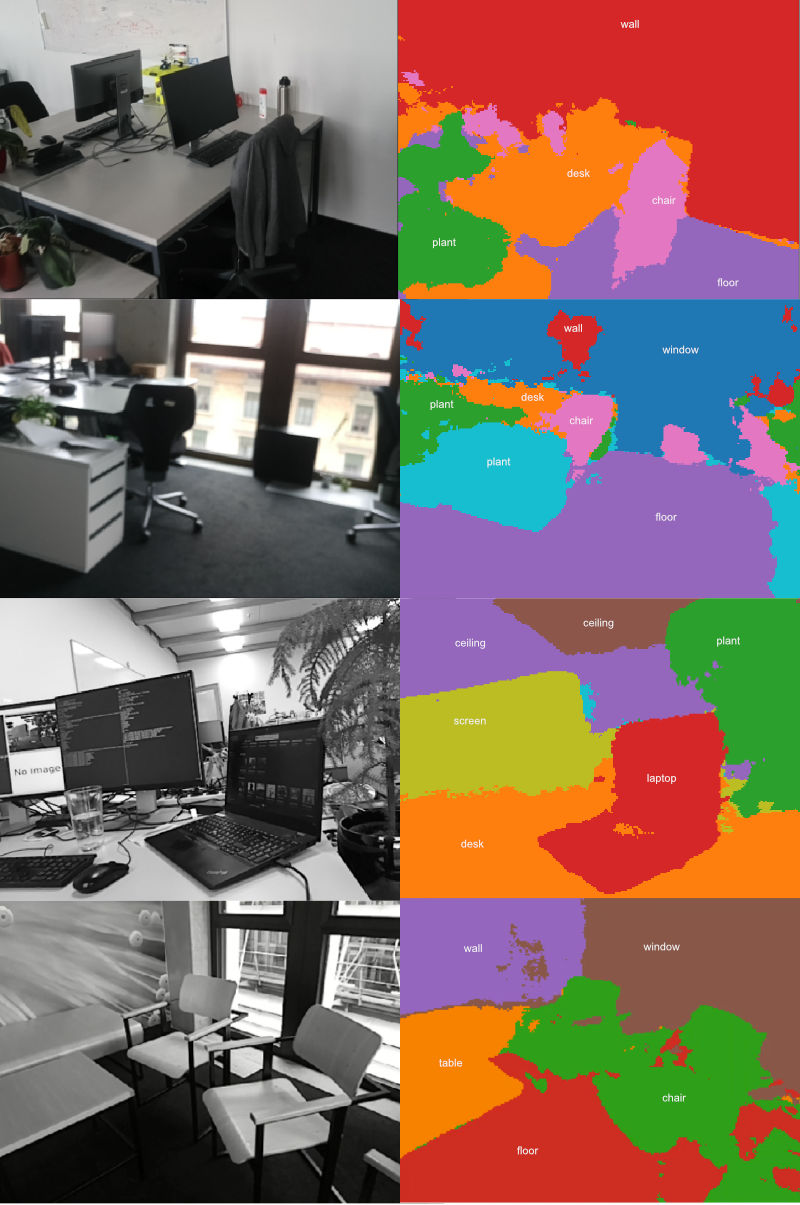}
    \caption{RGB renderings and semantic segmentation maps of our representation from our real-time experiment in an office environment given the prompts shown below the images.}
    \label{fig:real-time}
    \vspace{-1.5em}
\end{figure}

Figure \ref{fig:real-time} shows snapshots of a real-time experiment performed with a handheld camera in a regular office environment. The prompts used to produce the segmentation map are shown, but note that these can be changed at run-time to re-segment the scene. Figure \ref{fig:timeline} shows how quickly our representation is able to fit to a new scene when learned from scratch and integrating frames in real-time. After a dozen seconds, our method is able to produce good segmentation maps and scene reconstructions.

\subsection{Query Performance}

We time the latency and throughput of queries performed with our implementation on an Nvidia RTX $3070$ GPU. 3D semantic and density point queries can be performed at over $7$ million lookups per second with a latency of less than $10$ milliseconds. 2D ray queries can be rendered and segmented at roughly $\num{30 000}$ pixels per second using $256$ samples per ray, but this can be adjusted to to suit the desired fidelity.

\section{DISCUSSION AND CONCLUSIONS}
While the results obtained are an encouraging first step towards open-set 3D semantic segmentation there are still many open questions to improve such approaches, some of which we discuss in the following.

Currently, the largest factor limiting segmentation performance is the quality of the vision-language features. While LSeg uses natural language features from CLIP trained on a very large dataset, the visual encoder is trained on the small closed-set ADE20K dataset. If we were able to compute dense pixel-aligned visual-language features from open-set web scraped data without requiring any human annotations, we believe that results could eventually surpass supervised learning methods. \cite{ranasinghe2022perceptual} presented some promising initial results on learning pixel aligned features without using segmentation masks or other expert annotations. 

In real-time experiments, our system relied on poses coming from a SLAM system. If many bad poses are computed by the SLAM system, the 3D representation could become corrupted by bad updates. Possible solutions include treating the sparse SLAM poses as initial guesses and optimizing the poses jointly with scene geometry, as in \cite{sucar2021imap, zhu2022nice}, or bad poses could be filtered out by analyzing the photometric or geometric error across frames. 

In robotics, downstream modules, such as motion planners and high-level planning systems, might benefit from a more explicit and principled representation of geometry than what we presented in this paper. For example, signed distance function based approaches \cite{wang2021neus} might provide better surface and occupancy reconstruction and have other favorable properties, such as the ability to compute the normal of a surface by differentiating through the distance function. For the time being, our method is limited to static scenes. Dealing with moving objects within scenes remains an open problem, but promising recent research \cite{kong2023vmap} suggests that extending neural implicit representations to dynamic scenes might be feasible.

% \section{CONCLUSIONS}
To conclude, we proposed a volumetric neural representation which is able to jointly learn geometry, radiance, and semantic feature information of a scene. We have shown that by using dense pixel-aligned vision-language features, our resulting learned representation can be used to volumetrically segment scenes into, at run-time, user defined categories. We have also shown how the representation can be used to produce dense 2D segmentation maps for queried viewpoints. Experiments on the ScanNet dataset showed competitive performance and our real-world experiments demonstrate that the method could be run onboard a robotic system.

%\addtolength{\textheight}{-1cm}   % This command serves to balance the column lengths

%%%%%%%%%%%%%%%%%%%%%%%%%%%%%%%%%%%%%%%%%%%%%%%%%%%%%%%%%%%%%%%%%%%%%%%%%%%%%%%%

%%%%%%%%%%%%%%%%%%%%%%%%%%%%%%%%%%%%%%%%%%%%%%%%%%%%%%%%%%%%%%%%%%%%%%%%%%%%%%%%

%%%%%%%%%%%%%%%%%%%%%%%%%%%%%%%%%%%%%%%%%%%%%%%%%%%%%%%%%%%%%%%%%%%%%%%%%%%%%%%%

\bibliography{references}
\bibliographystyle{ieeetr}
% \printbibliography

\end{document}